\documentclass[final,3p,times]{elsarticle}
\PassOptionsToPackage{numbers, compress}{natbib}
\usepackage[utf8]{inputenc} 
\usepackage[T1]{fontenc}    
\usepackage{hyperref}       
\usepackage{url}            
\usepackage{booktabs}       
\usepackage{amsfonts}       
\usepackage{nicefrac}       
\usepackage{microtype}      
\usepackage{graphicx}
\usepackage{lineno}
\usepackage{amsmath}
\usepackage{algorithm}
\usepackage{algorithmic}
\usepackage{subfigure}
\DeclareMathOperator*{\argmin}{arg\,min}
\journal{}
\begin{document}
\begin{frontmatter}
\title{ Convolutional Network for Attribute-driven and Identity-preserving Human Face Generation}

%

\author[1]{Mu Li}
\author[2]{Wangmeng Zuo}
\author[1]{David Zhang}
\address[1]{Department of Computing, The Hong Kong Polytechnic University, Hong Kong }
\address[2]{School of Computer Science and Technology, Harbin Institute of Technology, Harbin 150001, China}


\begin{abstract}
   This paper focuses on the problem of generating human face pictures from specific attributes. The existing CNN-based face generation models, however, either ignore the identity of the generated face or fail to preserve the identity of the reference face image. Here we address this problem from the view of optimization, and suggest an optimization model to generate human face with the given attributes while keeping the identity of the reference image. The attributes can be obtained from the attribute-guided image or by tuning the attribute features of the reference image. With the deep convolutional network "VGG-Face", the loss is defined on the convolutional feature maps. We then apply the gradient decent algorithm to solve this optimization problem. The results validate the effectiveness of our method for attribute driven and identity-preserving face generation.
\end{abstract}
\end{frontmatter}
\section{Introduction}

Deep convolutional networks (CNNs) have shown great success in versatile high level vision problems such as image classification~\citep{krizhevsky2012imagenet}, face recognition~\citep{Parkhi15}, object detection~\citep{girshick2014rich}, image captioning, and visual question answering (VQA)~\citep{VQA}. CNNs have exhibited their remarkable power in understanding images at a fine level of granularity (e.g., pixel level), and can be adopted to generate pixel-level results from high level description, reference images or degraded observations. By far, CNNs have been successfully applied to semantic segmentation, image generation~\citep{kulkarni2015deep,gauthier2014conditional,dosovitskiy2015learning,yan2015attribute2image}, and some low level vision problems, including image denoising~\citep{xie2012image}, super-resolution~\citep{dong2014learning,johnson2016perceptual}, and image colorization~\citep{zhu2015learning}.


Several CNN-based models have been developed for human face generation with attributes. \citet{kulkarni2015deep} proposed deep convolution inverse graphic network (DG-IGN) which learns the latent graphic code from pictures and generates human faces with different pose and light by modifying the graphic code. However, this model needs to be trained by a large number of faces of a single person from different light and pose. These training face images are generated by 3D human face model. Gauthier~\citep{gauthier2014conditional} developed a conditional generative adversary network (cGAN) that tries to generate images from a noise distribution and conditional attributes. This model is trained through alternatively optimizing a generator and a discriminator. The generator is used to generate images being close to the natural images, while the discriminator is used to distinguish the generated images from the natural ones. \citet{yan2015attribute2image} developed an attribute-conditioned deep variational auto-encoder which abstracts the latent variables from a reference image with a recognition model and combined them with attributes to generate an image with given attributes by a generative model.

Unfortunately, the identity of the generated face is not emphasized in the existing CNN-based face generation models. Actually, in many graphics applications there are increasing requirements on generating face images for a specific identity. The problem can be formulated as transferring the given attributes while keeping the identity of the reference image. When the attributes are obtained from the attribute-guided image, it is also named as facial avatars. Even great advances have been achieved in facial avatars, most methods are based on 3D model and are physically unrealistic. Therefore, the development of CNN-based method can not only provide a new viewpoint for attribute driven and identity-preserving face generation, but also offers some new insight on the understanding and use of CNN.

In this paper, we consider face generation as an optimization problem defined on a pre-learned deep CNN model. Here we adopt the VGG-Face~\citep{Parkhi15} network which is trained on a very large scale dataset for face recognition. The optimization model involves an attribute term to transfer the given attributes to the generated face, an identity term to impose the identity constraint on the generated face, and a regularization term to generate smooth and sharp image. To solve the optimization problem, we apply gradient decent algorithm through the backpropgate of VGG-Face to generate face from a blank image. Moreover, to alleviate color mismatching with the reference, a color transfer is further learned to transform the generated image into the color space of the reference face. Furthermore, attribute mask is introduced to improve the visual quality of the generated image.

We evaluate the model on the Labeled Faces in the Wild (LFW)~\citep{LFWTech} database. In LFW, 73 visual attributes are generated by the attribute classifiers for face verification~\citep{kumar2009attribute}. Although there are some outliers for each class, these attributes and the corresponding faces are still sufficient to model the attribute term. We also analyze the choice of the convolutional layers for the attribute and the identity terms. The experiments demonstrate that our model can generate faces with given attributes while keeping the identity same with reference face. The generated faces are realistic and clearly.

\section{Related works}
As a classic topic in machine learning, generative model has been researched for a long time. A wide variety of deep learning approaches involve generative parametric models. The Restricted Boltzmann machine (RBM)~\citep{krizhevsky2010factored}, Deep Boltzmann machines~\citep{salakhutdinov2009deep} and Denoising auto-encoders~\citep{vincent2008extracting} are used for learning the image representation and reconstructing the image from it.

Recently, some interesting generative model has been proposed. \citet{goodfellow2014generative} proposed the Generative Adversarial Nets (GANs) to generate images from a noise distribution. In the GANs, a discriminator and a generator are alternatively trained as a adversarial competition game. The generator aims to generate images following the data distribution, while the discriminator attempts to distinguish between the generated images and the training data. \citet{kingma2013auto} proposed a stochastic variational auto-encoder from the view of the directed graphical model. A recognition network is used to approximates the intractable posterior with Gaussian latent variables. \citet{gregor2015draw} developed a recurrent auto-encoder with attention mechanism (DRAW) to generate images via a trajectory of patches.  All of these models generate images with a neural network trained with certain loss function. Different with them, our work model the generation problem from the the view of optimization.

Some 3D attribute transfer works are also related to our model. Most research on 3D attribute transfer (avatars)~\citep{kholgade2011content} aim to transfer facial attributes of human collected by sensors to an animator character. \citet{vlasic2005face} proposed a multilineal models to automatically transform the facial expression from person to person. This model is also created from 3D scans. \citet{suwajanakorn2015makes} developed a 3D face transfer system from a large photo collection. This system is a novel combination of 3D face reconstruction, tracking, alignment and multi-texture modeling. Different from these 3D attribute transfer methods, our model is based on 2D images and does not need to construct a 3D model.

Our work is also related to the works of visualizing and exploring the properties of CNN. \citet{mahendran2015understanding} proposed a framework to reconstruct images from HOG, SIFT and CNN representations and visualize them. \citet{simonyan2013deep} reconstruct images through maximizing a class score build on the output of ConvNet. \citet{gatys2015style} created images with artistic style by combining content and style losses. All these models view the image reconstruction problem as the optimization problem with loss function built on the representation. Compared with these works, we introduce an identity term and an attribute term built on VGG-Face for attribute-driven identity preserving generation of face images. Our  model not only keep the whole content of the image invariant but also modify the details of the image for attribute transfer. Moreover, spatial mask and color transform are further exploited to improve face generation quality.

\section{Face generation for attribute transfer}
In this section, we first define the perceptual loss in~\ref{s2_1}. Then our model is formulated as an optimization problem with the perceptual loss in ~\ref{s2_2}. Finally, we proposed two techniques to improve the quality of the generated image in ~\ref{s2_3}.

\subsection{Perceptual loss}\label{s2_1}
The perceptual loss is used to evaluate the semantic similarity between images in terms of attribute and identity. Instead of forcing two images to be exactly the same in pixel level, we require the content of two images to be similar. Since images with similar contents should have similar CNN features, the squared error loss between the CNN feature representations are adopted to define the perceptual loss.

We take use of the VGG-Face to model this semantic perceptual loss. VGG-Face introduced by Parkhi et al.~\citep{Parkhi15} is trained on a very large scale face dataset and has been shown to yield excellent performance on face recognition problem. There are over 40 layers in VGG-Face, we only exploit the first 5 convolution layers which could extract some common feature representation from the images. Deeper convolutional layers are left out due to their poor performance on image reconstruction.

Denote by $\phi$ the VGG-Face network. $\phi_l(I)$ is the feature map of the $l$th convolutional layer with respect to the input image $I$. $C_l$, $H_l$ and $W_l$ represent the channels, height and width of the feature map, respectively. The semantic perceptual loss between two images $I$ and $\hat{I}$ on the $l$th convolutional layer is defined as the squared-error loss between the two feature representations
\begin{equation}\label{eq1}
 \ell_{content}^{\phi,l}(\hat{I},I)=\frac{1}{2C_lH_lW_l}\|\phi_l(\hat{I})-\phi_l(I)\|_F^2
\end{equation}
Denote by $F_{i,j,k}^l(I)$ the activation value in the position $(j,k)$ of the $i$th channel of $\phi_l(I)$. The derivative of the loss function with respect to the feature maps in the $l$th convolution layer is calculated as
\begin{equation}\label{eq2}
 \rm \frac{\partial\ell_{content}^{\phi,l}(\hat{I},I)}{\partial F_{i,j,k}^l(\hat{I})}= \left\{ \begin{array}{ll}
 \frac{1}{C_lH_lW_l} (F_{i,j,k}^l(\hat{I}) -F_{i,j,k}^l(I)) &if \quad F_{i,j,k}^l(\hat{I})\geq0\\
 0 &if \quad F_{i,j,k}^l(\hat{I})<0 \end{array}\right.
\end{equation}
Then the gradient with respect to the image $\hat{I}$ can be computed by using the standard back-propagation process of the VGG-Face network.
\subsection{Model and algorithm}\label{s2_2}
\begin{figure}
  \centering
  \includegraphics[width=4in]{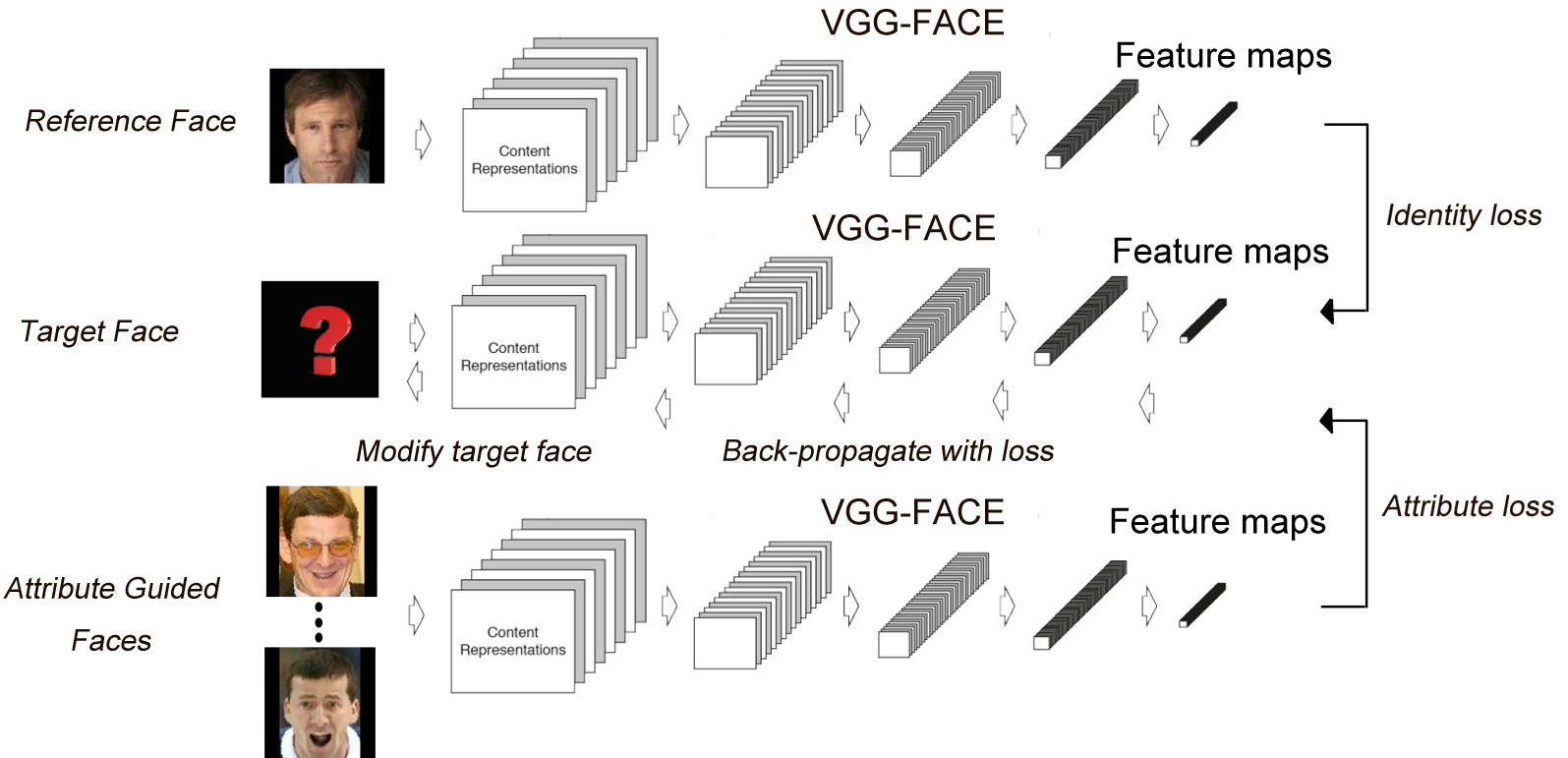}
  \caption{The framework of our model.}\label{networks}
\end{figure}

The goal of our work is to generate human face from given attributes and keep the identity. We model the generation problem as an optimization problem. On the one hand, the target face is required to be satisfied with the given attributes by minimizing an attribute loss. On the other hand, the target face should keep the identity through minimizing an identity loss.  Figure~\ref{networks} gives an illustration of the framework of our model.

\textbf{Attribute loss.}\quad The attribute loss is modeled on a set of images that contains the given attributes extracted from the LFW dataset. Assume that there is a set of images in the same pose with the same attributes given by human. The other attributes of these images are independently sampled from the corresponding attribute distribution. If the size of the set is large enough, the joint use of these face images will be helpful in enhancing the given attributes while suppressing the other attributes. Forcing the target face to be similar to the average face is a good solution. However, due to the number of faces in LFW is limited, it is difficult to get such large enough subset. Instead, we model the attribute loss on a small subset of LFW. This subset contains images with the given attributes. Meanwhile, images in it are similar to the reference face. Denote by this set the guided set. Then the attribute loss is defined as a weighted semantic perceptual loss between the target face and all the faces from the guided set.

To model the attribute loss, we should extract the attribute subset first. According to the attribute labels of LFW~\citep{kumar2009attribute}, a set of images containing the given attributes is first extracted. Then, the guided set which are closed to the reference faces should be further extracted from these images. We evaluate the similarity between images with a similarity measure. Instead of just comparing two faces in pixel level by simply calculating a squared error, we model this similarity measure from two aspects, i.e., pose and content.

For the pose similarity, we calculate the square loss between 68 landmarks~\citep{zhu2014transferring} extracted from the each face image. Denote by $px_i^j$ and $py_i^j$ the $x$-coordinate and the $y$-coordinate of the $j$th landmark of image $I_i$, the pose distance is defined as:
\begin{equation}\label{eq3}
  D_p(I_i,I_j) = \sum_{k=1}^{68}(px_i^k-px_j^k)^2+(py_i^k-py_j^k)^2
\end{equation}

For the content distance, we just take use of the sematic perceptual loss introduced in~\ref{s2_1}.

Denote by $\mathcal{D}(attributes)=\{I_1,I_2,\ldots,I_m\}$ the subset extracted from the LFW which contains the given $attributes$, where $I_i$ is the $i$th image in this set, $m$ is the size of this set. The similarity measure between the image $I_i$ from the set and the reference face $I_r$ is defined as:

\begin{equation}\label{eq4}
  D_a(I_i,I_r)=(1-\alpha)\frac{D_p(I_i,I_r)}{\sum_{k=1}^m D_p(I_k,I_r)}+\alpha \frac{\ell_{content}^{\phi,l}(I_i,I_r)}{\sum_{k=1}^m \ell_{content}^{\phi,l}(I_k,I_r)}
\end{equation}
where $\alpha$ is a tradeoff parameter between 0 and 1. We use Eq.~\eqref{eq4} to calculate the similarity distance between the reference face and all the images in $\mathcal{D}(attributes)$, and sort these images according to the distance in ascending order. The first $k$ images are selected to form the guided set $\mathcal{G}(attributes)=\{s_1,s_2,\ldots,s_k\}$, where $s_i$ represents a face image. The attribute loss is based on the set $\mathcal{G}(attributes)$ as follows.
\begin{equation}\label{eq5}
  \ell_{attr}(\hat{I})=\sum_{i=1}^k w_i\ell_{content}^{\phi,l}(\hat{I},s_i)
\end{equation}
 where $\hat{I}$ is the target face to be generated, $w_i$ is the weight to control the contribution of the face $s_i$ with $\sum_{i=1}^k w_i =1$.

\textbf{Identity loss.}\quad The identity loss is used to encourage the target face to keep the same identity with the reference face. Instead of requiring all the details of two face to be exactly same, the identity loss is also based on the semantic perceptual loss mentioned in~\ref{s2_1}. This loss is used to force the target face $\hat{I}$ and the reference face $I_r$ to be similar on the whole. The identity loss is defined as:
\begin{equation}\label{eq6}
  \ell_{id}(\hat{I})=\ell_{content}^{\phi,l}(\hat{I},I_r)
\end{equation}
\textbf{Regularization.}\quad In order to encourage the spatial smoothness of the generated face $\hat{I}$, we take use of the total variation (TV) regulariser $\ell_{TV}(\hat{I})$. This regulariser is also used to reconstruct image from feature representations~\citep{mahendran2015understanding} and transform style~\citep{johnson2016perceptual}.

\textbf{Model.} Our goal is to generate a face image from the reference face with given attributes without changing the identity. This can be modeled as an optimization problem with the attribute loss term and the identity loss term. The target face image is generated by minimizing the attribute and the identity loss terms simultaneously. In addition, the TV regulariser is added to encourage the smoothness of the generated face. Thus, our model is defined as:
\begin{equation}\label{eq7}
\hat{I}= \argmin_{\hat{I}} \;\ell_{attr}(\hat{I})+\lambda\ell_{id}(\hat{I})+\gamma \ell_{TV}(\hat{I})
\end{equation}
where $\hat{I}$ is the target face to be generated. $\gamma$ is the tradeoff parameter for the TV regulariser. $\lambda$ is the tradeoff parameter for the identity loss.

\textbf{Algorithm.}\quad  As shown in Algorithm~\ref{alg1}, we apply gradient decent algorithm on the optimization problem. The gradient of attribute loss and identity loss with respect to $\hat{I}$ can be easily got by following the derivative process of the semantic perceptual loss.
\begin{algorithm}[!tbp]
\caption{Attribute transfer for identity preserving face generation}\label{alg1}
\begin{algorithmic}[1]
\renewcommand{\algorithmicrequire}{\textbf{Input:}}
\renewcommand{\algorithmicensure}{\textbf{End}}
\REQUIRE Face image set $\mathcal{T}$, Attribute label set $Y$, Given attribute $att$, Reference face $I_r$.
\renewcommand{\algorithmicrequire}{\textbf{Output:}}
\renewcommand{\algorithmicensure}{\textbf{End}}
\REQUIRE The generated face $\hat{I}$
\STATE
Extract a subset $D(attr)$ from $\mathcal{T}$ whose attribute label satisfied with the given attributes $att$.
\STATE
Calculate the similarity distance with Eq.~\eqref{eq4} between all the images in $D(attr)$ and the reference face $I_r$.
\STATE
Extract the first $k$ faces in $D(attr)$ that has the smallest similarity distance.
\WHILE {not converged}
\STATE
Calculate $G(\hat{I})$ the gradient of the optimization problem with respect to the target face $\hat{I}$ in Eq.~\eqref{eq7} with the $k$ faces extracted in step 3.
\STATE
$\hat{I} \leftarrow \hat{I}-aG(\hat{I})$, $a$ is the learning rate.
\ENDWHILE
\STATE
Return $\hat{I}$
\end{algorithmic}
\end{algorithm}

\begin{figure}
  \centering
  \includegraphics[width=4in]{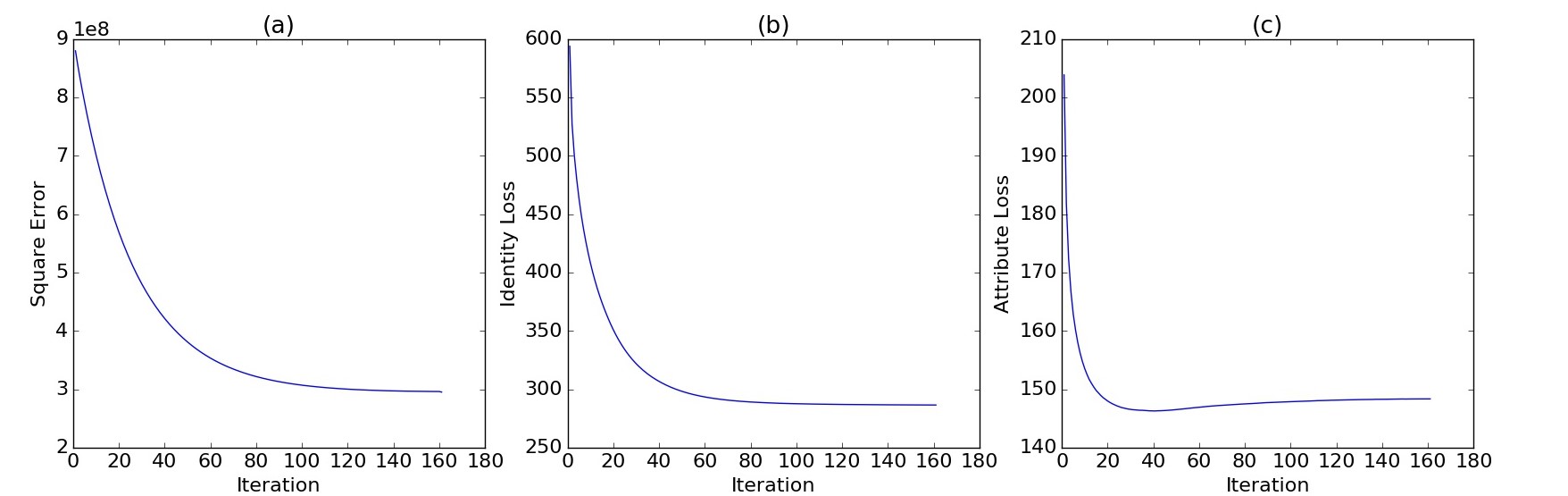}
  \caption{Evaluation of the Algorithm~\ref{alg1}. (a) shows curve of the square difference error between the generated face and the reference face with the iteration increasing. (b) shows curve of the identity loss. (c) shows the curve of the attribute loss.}\label{loss}
\end{figure}
As shown in Figure~\ref{loss}, in experiment, the identity loss and the attribute loss are first reduced and then keep roughly unchanged along with the increase of iterations. It shows that our algorithm works well for this optimization problem. For the squared error between the generated face and the reference face, it decreases to certain threshold and then keep unchanged. This shows the generated face is not exactly the same with the reference due to some attributes have been modified.

\subsection{Improvements}\label{s2_3}
\begin{figure}
\centering
\subfigure[Generate face with mask]{
\label{mask} 
\includegraphics[width=2in]{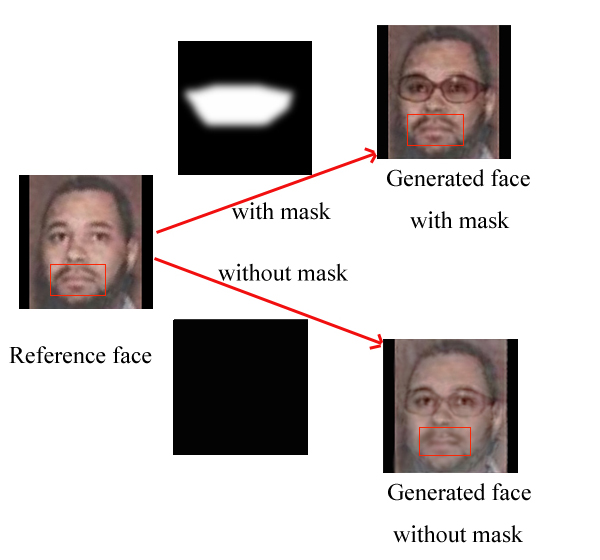}}
\hspace{0in}
\subfigure[Color transformation]{
\label{color} 
\includegraphics[width=2in]{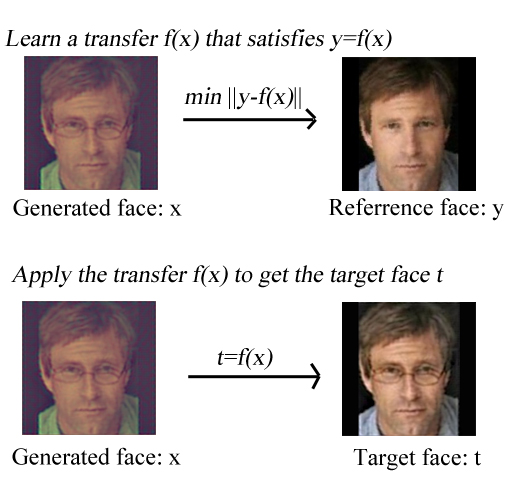}}
\caption{Illustration of the two improvement technics.}
\label{tricks} 
\end{figure}

\textbf{Adding Spatial Masks for Attributes}.\quad As mentioned in~\ref{s2_2}, we take use of the face images with the given attributes that are similar with the reference to model the attribute loss. However, as the size of the dataset is limited, for some reference face, it is hard to find similar faces with given attributes. Those dissimilar faces would modify some attributes that are not given by the users, which will further decrease the quality of the generated face. To handle this problem, we introduce attribute masks to restrict the effective area of the attribute loss. Particularly, the attribute loss can only be applied on the spatial area where the mask value is larger than $0$. Figure~\ref{mask} shows that adding mask could help to keep the other attributes unchanged. In Figure~\ref{mask}, we generate a face with the attribute "with glasses" from the reference face. Before adding mask, the generated face also changes the attributes on mouth and mustache, while the one generated with the mask keep all the other attributes unchanged. Here, all the attributes mask are generated from the 68 landmarks extracted from each face. For each attribute, we associate it with certain landmarks and generate a convex hull for these landmarks. After that, we expand the convex hull with a large margin to get the attribute masks. Though the masks created by this technique do not exactly bound the attributes, we found that it still improve the quality of generated faces. If more than one attribute is given, a union of all the attribute masks is calculated. With the introduced attribute mask, the attribute loss in Eq.~\eqref{eq5} is modified as:
\begin{equation}\label{eq8}
  \ell_{attr}(\hat{I},M)=\sum_{i=1}^k w_i\ell_{content}^{\phi,l}(\hat{I}\bigotimes M,s_i)
\end{equation}
where $M$ is the attribute mask, and $\bigotimes$ represents element-wise multiplication in pixel level. Denote by $\hat{I}_{ijk}$ the pixel value at position $(i,j)$ on channel $k$ of the target face $\hat{I}$, and $M_{ijk}$ the corresponding mask value. The gradient of  $\ell_{attr}(\hat{I},M)$ with respect to $\hat{I}$ should be calculated as:
\begin{equation}\label{eq9}
 \frac{\partial \ell_{attr}(\hat{I},M)}{\partial \hat{I}_{ijk}}= \left\{ \begin{array}{ll}
 \frac{\partial \ell_{attr}(\hat{I})}{\partial \hat{I}_{ijk}} &if \quad M_{ijk}>0\\
 0 &if \quad M_{ijk}\leq0 \end{array}\right.
\end{equation}
\textbf{Transforming the Color of Generated Face.}\quad The original face generated from our model suffers from poor color quality which makes it look unrealistic. The main reason is that color information is lost after several convolution and pooling operations in VGG-Face network. In order to make our generated face looks more realistic, we designed a linear color transform to transform the color of the generated face to the color space of the reference face. Here, all the transform operation is under the YCbCr color spaces. Denote $\mathbf{x}=(\mathbf{x}_y; \mathbf{x}_{cb}; \mathbf{x}_{cr})$ as a pixel in YCbCr space, the transform function is defined as Eq.~\eqref{eq10}. Here, $\mathbf{A}$ is a $3\times3$ matrix. For each pixel in the input image, we define the transform operation as,
 \begin{equation}\label{eq10}
 f(\mathbf{x}) = \mathbf{A} \mathbf{x}
 \end{equation}
As shown in Figure~\ref{color}, the transform matrix $\mathbf{A}$ is learnt by minimizing $||y-f(t)||_F^2$. Here, $y$ is the reference face, $t$ is the target face generated. Instead of applying all the corresponding pixels between $t$ and $y$ to solve the matrix $\mathbf{A}$, we sample 1,000 pixels from the central part of $t$ to generate a $3\times1000$ matrix $\mathbf{X^\prime}$ and the corresponding pixels from $y$ to generate the matrix $\mathbf{Y^\prime}$. Then we calculate $\mathbf{A}$ as $\mathbf{A}=(\mathbf{X^\prime}^T\mathbf{X^\prime})^{-1}\mathbf{X^\prime}\mathbf{Y^\prime}$.

\section{Experiments}
Our experiments are conducted on the Labeled Faces in the Wild (LFW) dataset which contains about 13,000 face images. All the faces are aligned with 68 landmarks~\citep{zhu2014transferring} extracted from each face and rescaled to the size of $224\times224$, which is the input size of VGG-Face. However, many faces in LFW are of low resolution. Rescaling them to $224\times224$ would make the face images blurry. In order to enlarge the dataset, we augment the dataset by flipping the images horizontally which will not change the visual attributes.

\subsection{Attribute based image generation}\label{s4_1}
We apply our attribute based face generation algorithm on LFW to generate face images with different attributes from the reference faces. All the reference face is randomly selected from the LFW dataset. In our experiments, $k$ is set as $5$. Since the attribute labels are generated by support vector machine trained in~\citep{kumar2009attribute}, about 10\% labels are not correct. In experiment, we leave out the faces with wrong attribute label and blurry visual attributes.
\begin{figure}
  \centering
  \includegraphics[width=4in]{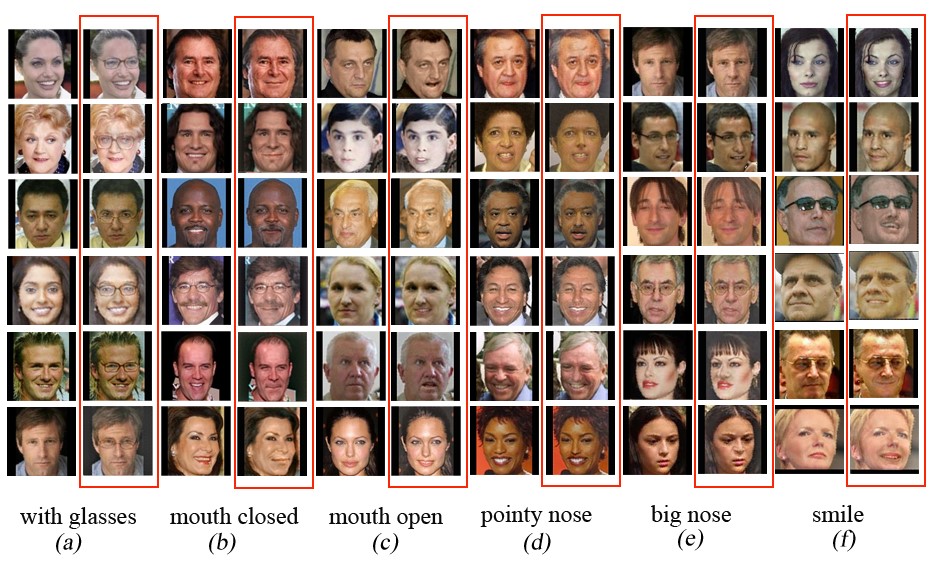}
  \caption{Images generated with different visual attributes. For each attribute, the left collum is the reference face and the right collum is the generated face.}\label{gimg}
\end{figure}

In our experiments, as shown in Figure~\ref{gimg}, we generate images with the attribute "with glasses", "mouth closed", "mouth open", "pointy nose", "big nose" and "smile", respectively. It shows that our model can achieve amazing results not only on the partial attribute such as "with glass", "mouth closed", "mouth open", "pointy nose" and "big nose" but also on the global attributes "smile". Our generation model can generate the faces with given attribute while preserving the identity.

\subsection{Generate images with given guided images}
\begin{figure}
  \centering
  \includegraphics[width=4in]{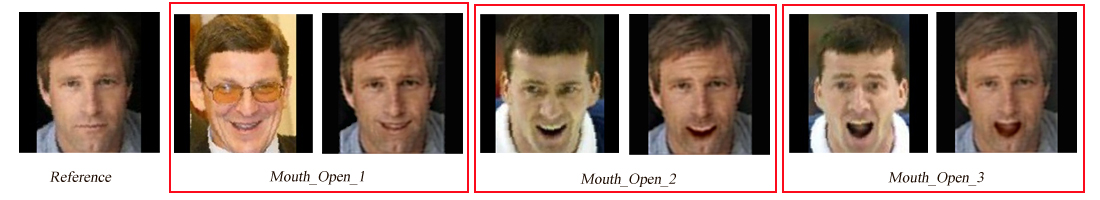}
  \caption{Faces generated from given guided images.}\label{guide}
\end{figure}
Our model can also be used to generate face images by borrowing visual attribute from given guided faces. In this experiment, we do some small modification for our model. Instead of extract the guided images from LFW dataset with the similarity distance, we provide the guided image directly. As shown in Figure~\ref{guide}, we give three guided faces. And three target faces are generated from the reference with the visual attribute of mouth transformed from the guided images. In Figure~\ref{guide}, the left image in each group is the guided face, and the right one is the generated face.

\subsection{Evaluate the quality generated images}
\subsubsection{Images generated from different VGG layers}
\begin{figure}
  \centering
  \includegraphics[width=3in]{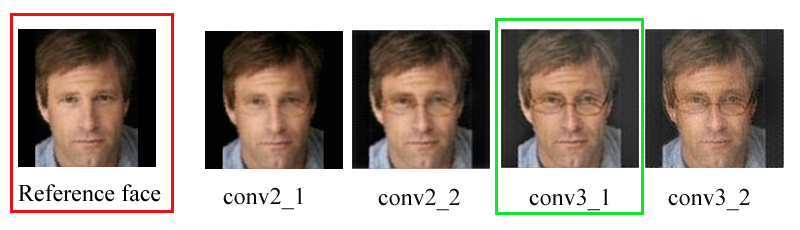}
  \caption{Faces generated from different convolutional layers.}\label{layers}
\end{figure}

In this experiment, we test our generation model with the loss function built on different convolutional layers of VGG-Face. All the parameters used to generate the faces are exactly same except for the layer chosen to model the loss function.  Figure~\ref{layers} shows the face images with the attribute "with glass" generated from different convolutional layers. Table~\ref{table:layer} shows the square difference error between the generated faces and the reference face. As we can see, face generated from "\texttt{conv2\_1}" has the highest image quality but the most inconspicuous given attribute, while face generated from "\texttt{conv3\_2}" has the poorest image quality but conspicuous attribute. From the results, we can conclude that semantic perceptual loss modeled on lower convolutional layer could reconstruct the reference face more accurately but has lower tolerance of attributes modification, while perceptual loss on higher convolutional layer is just the opposite. For our image generation experiment conducted in~\ref{s4_1}, we take the tradeoff between the image quality and the salience of the attribute and choose to use "\texttt{conv3\_1}" to model the loss function.

\begin{table}[htb]
\scriptsize
\begin{center}
\begin{tabular}{l|c|c|c|c}
\hline
Convolution layer& \texttt{conv2\_1} ($3$rd) & \texttt{conv2\_2} ($4$th) & \texttt{conv3\_1} ($5$th) & \texttt{conv3\_2} ($6$th)  \\
\hline
Square Error ($\times10^8$)	& 1.65 &	1.73 & 2.95 & 3.44 \\
\hline
\end{tabular}
\end{center}
\caption{Square Error of the faces generated from different convolutional layers}
\label{table:layer}
\end{table}
\subsubsection{Influence of the TV regulariser $\ell_{TV}$}
\begin{figure}
  \centering
  \includegraphics[width=2.5in]{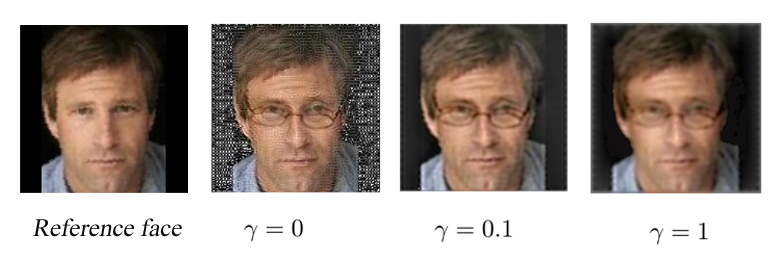}
  \caption{Faces generated with different $\gamma$ values.}\label{tv}
\end{figure}
The TV regularizer has an important influence on the quality of the generated images. The gradient used to generate the images is back-propagated from a deep convolutional layer. During this process, some information may be lost or modified. Thus, the generative gradient consists of lots of noise. Without the TV regularizer, the generated images will contains noise and mosaics which will decrease the quality of the images. In Figure~\ref{tv}, we generate face images from the attribute "with glasses" with different $\gamma$ values to control the contribution of the TV regularizer in the loss function. As we can see, when $\gamma=0$, the generated face is noisy and in low quality; while when $\gamma=1$, the generated face is over-smooth and the outline of the glasses is blurry. Thus, the choice of $\gamma$ does make influence on the quality of generated images.

\section{Conclusion}
In this paper, we model the attribute driven face generation problem as an optimization problem with semantic perceptual loss. Spatial mask and color transform are further introduced to improve face generation quality. The gradient decent algorithm is then used to solve the optimization problem defined on the VGG-face network. Experiments show that our model can generate realistic faces with given attributes while preserving the identity of reference face. In the future, we will incorporate with the CNN learning to improve the efficiency and effectiveness of our approach.

\small
\bibliographystyle{model1-num-names}
\bibliography{database}

\end{document}